\documentclass[mono]{riverk}
\usepackage{hyperref}
\usepackage{rivps}
\usepackage{graphicx}
\usepackage{amsmath}
\usepackage{tabularx}
\usepackage{polyglossia}
\setmainlanguage{english}
\setotherlanguages{arabic,hindi,sanskrit,greek,thai}

\usepackage{makeidx}
\makeindex

\begin{document}
\frontmatter

\author{Praveenkumar Katwe, RakeshChandra Balabantaray, Kaliprasad Vittala}
\title{Bridging the Data Gap: Creating a Hindi Text Summarization Dataset from the English XSUM}
\maketitle

\tableofcontents

\chapter*{Preface}

Creating a dataset in Hindi for XSUM, a task focused on text summarization, represents a pivotal step towards bridging linguistic gaps in natural language processing (NLP) and making state-of-the-art technologies accessible and relevant to a wider audience. This chapter delves into the multifaceted process of dataset creation, specifically tailored to the needs and nuances of the Hindi language, a rich and complex linguistic system spoken by hundreds of millions of people.

The journey of creating such a dataset is both challenging and rewarding. It involves careful consideration of linguistic diversity, cultural nuances, and the technical requirements of text summarization models. This chapter aims to guide readers through the intricacies of this process, from the initial planning stages to the final execution, highlighting the importance of linguistic inclusivity in the development of NLP technologies.

We begin by exploring the motivations behind creating a Hindi dataset for XSUM, emphasizing the need to extend the benefits of text summarization to non-English languages. The chapter then outlines the steps involved in dataset creation, including data sourcing, annotation, and preprocessing, with a focus on handling the unique characteristics of Hindi. Special attention is given to the challenges encountered during this process, such as dealing with code-mixed language, partial translations and ensuring the representation of diverse voices within the dataset.

As we navigate through these topics, the chapter also sheds light on the methodologies adopted to maintain the quality and reliability of the dataset. This includes strategies for annotation, leveraging expert knowledge, and employing automated tools to assist in the process. Furthermore, we discuss the ethical considerations inherent in dataset creation, such as privacy concerns, consent, and bias mitigation, which are crucial for building fair and unbiased AI systems.

The culmination of this chapter offers insights into the potential impacts of the Hindi XSUM dataset on the landscape of NLP, not only in enhancing text summarization capabilities for Hindi texts but also in inspiring similar efforts for other languages. Through this endeavor, we aim to contribute to the democratization of AI, ensuring that the benefits of technological advancements are equitably distributed across linguistic boundaries.

In essence, this chapter serves as both a practical guide and a philosophical reflection on the significance of linguistic diversity in AI. It is a testament to the collective effort required to extend the frontiers of technology to accommodate the rich tapestry of human languages, starting with Hindi.

\mainmatter

\part{The Data Gap}
\chapter{Bridging the Data Gap: Creating a Hindi Text Summarization Dataset from the English XSUM}
\section{Abstract}
The advancement of natural language processing (NLP) and machine learning (ML) technologies has predominantly focused on resource-rich languages, notably English. This has led to a significant gap in dataset availability and quality for low-resource languages, like Hindi, especially in specialized tasks such as text summarization. Text summarization, a crucial NLP application, entails compressing a longer text into a brief, informative summary. The development of text summarization models depends heavily on extensive and varied datasets, yet their scarcity in low-resource languages impedes progress in diverse linguistic settings.Addressing this gap, our study not only proposes the development of a comprehensive text summarization dataset for Hindi, derived from the English Extreme Summarization (XSUM) dataset but also introduces a novel and cost effective approach for automating dataset creation for low-resource languages. This method leverages advanced translation and linguistic adaptation techniques and validation using the Crosslingual Optimized Metric for Evaluation of Translation (COMET), ensuring high fidelity in translation and contextual relevance with selective usage of existing LLMs and curation.The proposed Hindi dataset, a robust, translated version of XSUM, seeks to enhance research in Hindi text summarization, providing a direct resource and contributing to the broader understanding of cross-lingual NLP challenges. This dataset will encompass a variety of topics and writing styles, mirroring the diversity of the original XSUM dataset. Such an approach ensures that the Hindi dataset is not only extensive but also representative of different textual complexities and themes.In conclusion, creating a Hindi text summarization dataset from English XSUM, complemented by an innovative automated dataset building approach using COMET for translation validation, marks a substantial stride towards democratizing NLP research and applications by reducing cost. By offering a task-specific dataset for a low-resource language, this initiative fosters more nuanced and culturally relevant NLP models and stimulates further research in text summarization, particularly for languages historically underserved in computational linguistics.
\section{Introduction}
\subsection{Overview of text summarization}
Text summarization is a crucial task in natural language processing (NLP) aimed at reducing the length and complexity of textual documents, ensuring that the essence and most critical information are retained. This process helps in managing the overwhelming amount of information available in digital texts, making it easier for users to quickly grasp the main points without reading the entire content. There are two primary approaches to text summarization: extractive and abstractive. Extractive summarization works by identifying and compiling key sentences or phrases directly from the source text without altering the original text, thus creating a summary that is a subset of the original. Abstractive summarization, on the other hand, involves generating new sentences that capture the core ideas of the text, often resulting in more natural and cohesive summaries that may not necessarily use the same phrasing as the source document.\cite{allahyari2017text}.

The advancement of deep learning techniques, particularly those involving models like BERT (Bidirectional Encoder Representations from Transformers) and GPT (Generative Pre-trained Transformer), has significantly enhanced the capabilities of text summarization systems. These models are trained on vast amounts of data and leverage complex architectures to understand context, semantics, and the nuances of language, enabling them to generate summaries that are both accurate and contextually relevant. Such technologies have found applications across various fields, including journalism, legal document analysis, and academic research, where quick synthesis of large texts is invaluable.\cite{Suleiman2020Deep}.

Furthermore, the development of these summarization models has spurred research into addressing challenges such as maintaining coherence in summaries, reducing bias, and ensuring the summaries are factually consistent with the source texts. The ongoing research and development in text summarization not only promise to enhance the efficiency and effectiveness of information retrieval and comprehension but also pave the way for innovative applications that can leverage summarized content for diverse purposes.\cite{liu2019text}

\subsection{Introduction to the XSUM}
The Extreme Summarization (XSUM) dataset stands as a distinctive and challenging benchmark in the domain of text summarization. Introduced by Narayan, Cohen, and Lapata in 2018, XSUM was designed to push the boundaries of summarization tasks by focusing on generating highly abstractive summaries. Unlike traditional summarization datasets that might encourage models to extract sentences or parts thereof directly from the text, XSUM aims to produce a single-sentence summary for each document, compelling models to truly synthesize and condense information. This dataset comprises over 200,000 articles from the British Broadcasting Corporation (BBC), spanning various topics and providing a rich source of diverse news content. The XSUM dataset challenges natural language processing models to capture the essence of an article in a concise manner, testing their ability to understand and reproduce content creatively and succinctly, thus making it a crucial resource for advancing abstractive summarization research. \cite{narayan2018don}
\subsection{Importance of language diversity in text summarization}
The importance of language diversity in text summarization is a critical aspect that reflects the evolving needs and challenges in natural language processing (NLP). Language diversity enriches NLP models by broadening their understanding and adaptability across different linguistic, cultural, and contextual spectrums, making these models more inclusive and effective on a global scale.

Language diversity in text summarization involves training models on datasets from a wide array of languages, which helps in capturing the rich, nuanced differences across linguistic features such as syntax, semantics, and pragmatics. This approach is essential not only for the generalization capabilities of NLP models but also for ensuring that AI technologies are accessible and useful to speakers of all languages. Tiedemann and Thottingal (2020) emphasize the significance of multilingual datasets in improving the performance of summarization models, highlighting how exposure to diverse linguistic structures enhances a model's ability to understand and generate summaries across languages.

Moreover, language diversity addresses the issue of bias in AI, ensuring that models do not overfit to the nuances of a single language, typically English, which has been predominantly represented in NLP research. Aida et al. (2021) discuss how language diversity can mitigate biases and foster fairness in AI applications by providing equal representation to underrepresented languages and dialects, thus democratizing the benefits of AI technologies.

Additionally, incorporating language diversity in text summarization tasks supports the preservation of linguistic heritage and promotes cultural understanding. By acknowledging and incorporating diverse languages, NLP technologies can play a pivotal role in documenting and revitalizing endangered languages, offering a digital platform for their study and preservation (Smith et al., 2022).

In conclusion, language diversity in text summarization is not just a technical necessity for improving model performance; it is a step towards ethical AI development, promoting inclusivity, fairness, and cultural preservation. As NLP continues to evolve, the integration of language diversity will be paramount in shaping globally aware and culturally competent AI systems.

\nocite{*}

\section{The Significance of Hindi in Natural Language Processing}

\subsection{Demographic and linguistic overview of Hindi}
Hindi, as one of the most widely spoken languages in the world, holds significant importance in the field of natural language processing (NLP). With over 340 million native speakers and more than 600 million speakers worldwide, Hindi stands as the primary lingua franca across northern India.\cite{campbell2008ethnologue} This demographic and linguistic prominence underscores the necessity for Hindi's inclusion in NLP research and development to cater to a vast and diverse population. Linguistically, Hindi is characterized by its use of the Devanagari script and exhibits a rich morphological structure, including the use of postpositions, gendered nouns, and a verb-final sentence structure, which presents unique challenges and opportunities for NLP applications \cite{ShuklaKumar2023}. The development of Hindi NLP tools and resources, such as morphological analyzers, part-of-speech taggers, and syntactic parsers, is crucial for advancing text analysis, machine translation, and speech recognition technologies for Hindi speakers. Additionally, the integration of Hindi into multilingual NLP models facilitates cross-lingual understanding and communication, further emphasizing the global significance of linguistic diversity in the digital age \cite{vinay2024}.
\subsection{Current Challenges and opportunities in Hindi language processing}
The researchers \cite{desai2021taxonomic} performed a comprehensive analysis of current state of hindi langage NLP system across tasks and had the below observations: 
• Many of developed tools/resources are not released in open domain in Hindi.So standardization of testing not possible.
• The language resource of wordnet(WN) still has scope of improvement.AS of today, English wordnet has 77194 more synsets then Hindi WN .Also there are lots of standard thesaurus and annotated corpus available for English.
• The MT was mostly from English to Hindi, but from Hindi to English not much effort found.
• For IR application, cross language IR improvement is needed for Hindi to English IR systems.
• The TS applications in Hindi are made mostly using extractive methods, abstractive techniques need to explored.
• There are many open issues like sarcasm detection, irony detection, humour identification,
stance detection etc. which are yet not fully explored in IL under sentiment classification.

There is scope for research on many of the task, however focusing on the Text Summarization,the major challenges being faced are : (i) Dataset availability (ii) Morphology analysis (iii)  Availability of NLP parsers. \cite{Kumar2021Study}

\section{Designing a Dataset for Hindi Text Summarization}

\subsection{Objectives and requirements of a Hindi dataset for XSUM}
The creation of a Hindi dataset for text summarization entails specific objectives and requirements to effectively address the linguistic and cultural nuances associated with Hindi language processing. The primary objective is to develop a comprehensive resource that enables the training and evaluation of text summarization models specifically tailored for Hindi, considering its syntactic and morphological features. Such a dataset must include a diverse range of texts covering various domains, such as news articles, academic papers, and literature, to ensure the models can handle different styles and contexts of Hindi text. Additionally, it is imperative to include annotated summaries that serve as references for both extractive and abstractive summarization tasks, facilitating the development of models capable of generating concise, coherent, and contextually relevant summaries in Hindi \cite{mamidala2021}

Moreover, the dataset should account for the linguistic diversity within Hindi, including dialectal variations and code-switching instances, to enhance the models' robustness and their ability to understand and summarize colloquial and non-standard forms of Hindi. Ensuring high-quality annotations is another critical requirement, involving the meticulous selection of annotators proficient in Hindi and trained in summarization techniques. This would help in achieving consistency and reliability in the annotated summaries, which is crucial for the effective training and benchmarking of summarization models. Furthermore, ethical considerations, such as the privacy of data sources and the representativeness of the content, should be addressed to promote fairness and mitigate bias in the resulting NLP tools.\cite{engproc2023059194}

\subsection{Considerations for dataset design}
Data quality and consistency are also paramount. This involves establishing clear guidelines for annotation to minimize subjectivity and variance among annotators, which can significantly affect the reliability of the dataset. Moreover, considering the linguistic and cultural nuances specific to the language of the text is vital, especially for languages with rich morphological features and diverse dialects \cite{hershcovichchallenges}. Additionally, ethical considerations such as consent for data use, privacy concerns, and the potential for bias in the dataset composition need careful attention to avoid perpetuating or introducing biases in summarization models developed using the dataset. Lastly, the size of the dataset is a practical consideration, as larger datasets typically provide more comprehensive training material but require more resources to collect and annotate \cite{liu2023responsible}.

\subsection{Overview of dataset structure and annotation guidelines}
The structure of datasets for abstractive text summarization and their annotation guidelines are crucial for the development of models that can generate coherent and concise summaries reflecting the essence of the original text. Typically, such datasets are structured to include pairs of full-text documents and their corresponding summaries. The source documents are selected to represent a wide array of topics and styles, ensuring that the dataset covers a diverse set of summarization challenges. Each document is paired with one or more reference summaries, which are created by human annotators trained in summarization techniques. These reference summaries are not mere extractions of sentences from the original text but are instead newly written texts that capture the main points in a condensed form, often involving paraphrasing, generalization, and the integration of information from across the document.

Annotation guidelines for abstractive summarization datasets emphasize the importance of capturing not only the factual content but also the tone, intent, and nuanced meanings of the original text.These guidelines also address issues of bias, ensuring that summaries do not introduce or propagate biases that were not present in the original text. Consistency in the application of these guidelines is maintained through rigorous training of annotators and the use of inter-annotator agreement measures to assess and improve annotation quality

\subsection{Data selection and Preparation}

The Extreme Summarization (XSUM) dataset is a pivotal resource in the text summarization field due to its unique focus on generating highly abstractive summaries. Unlike other datasets that might allow for or even favor extractive summarization strategies, where content is directly pulled from the source text, XSUM challenges models to produce a single-sentence summary that captures the essence of an entire article. This characteristic fosters the development and testing of NLP models on their ability to genuinely understand and distill complex information into concise, coherent summaries without relying on the source text's structure. This emphasis on high abstraction levels makes XSUM particularly important for advancing the capabilities of summarization technologies to produce more human-like summaries, pushing the boundaries of current AI's understanding and synthesis of natural language.

Compared to datasets like XLSum and CNN/DailyMail (CNN/DM), XSUM offers a more challenging benchmark for abstractive summarization. The CNN/DM dataset, for instance, is known for its suitability for extractive summarization tasks, given its structure that includes summaries often made up of sentences appearing verbatim in the text. Similarly, XLSum, while offering a multilingual perspective and diverse news sources, still presents summaries that can frequently be approached with extractive techniques due to their relatively longer and more detailed nature. In contrast, XSUM's requirement for single-sentence, highly abstractive summaries necessitates advanced understanding and generation capabilities from models, making it a more rigorous testbed for evaluating the state-of-the-art in abstractive summarization. Therefore, XSUM not only enriches the landscape of datasets available for summarization tasks but also sets a higher standard for model performance, driving innovation in the creation of more advanced, nuanced NLP systems 
\subsection{Structure of XSUM Articles}
The Extreme Summarization (XSUM) dataset is designed to facilitate the training and evaluation of models for highly abstractive text summarization. Structurally, the dataset consists of pairs of articles and single-sentence summaries. Each pair includes a full news article from the BBC and a corresponding summary that aims to capture the essence of the article in just one sentence. This design challenges models to comprehend and condense the article's main point, encouraging the development of advanced natural language understanding and generation capabilities.
Following is a structure of an article in XSUM data set.
1. **Article**: The body of a news article, which can range in length but typically contains several paragraphs of detailed information, including background, quotes, and various related points.
2. **Summary**: A single sentence that encapsulates the central idea or the most newsworthy aspect of the article. The summary is crafted to be highly abstractive, often requiring synthesis and generalization beyond simple extraction.

Sample Example

To illustrate, consider a hypothetical example based on the XSUM's structure:
- **id**: This is the unique identification number of the new article.

- **Article**: "The National Health Service has announced an increase in funding for mental health services across the UK. Over the next five years, an additional £2.3 billion will be allocated to support a range of initiatives aimed at improving access to psychological therapies, expanding community-based care, and enhancing treatment for young people with early signs of mental health issues. The government's decision comes in response to growing concerns about the rising rates of mental health problems and aims to address the long-standing underfunding in this sector. Health Secretary Jane Doe stated, 'This investment is a pivotal step towards reshaping mental health services and making sure these services are accessible to everyone, irrespective of where they live.'"

- **Summary**: "The UK's National Health Service will boost mental health funding by £2.3 billion to improve services and treatment nationwide."

This example demonstrates how the summary distills the key information from the article into a concise statement, highlighting the funding increase and its purpose without delving into the specifics and background details provided in the full text.

The XSUM dataset's emphasis on such high-level abstraction makes it a challenging and valuable resource for advancing text summarization research, pushing the boundaries of what automated systems can understand and generate from complex textual information. The articles within in the XSUM are divided into three categories. The dataset is typically divided into three subsets for the purposes of training, validation, and testing machine learning models.

\section{Preprocessing of the articles}
 The data set is converted to a json array of 25 articles and each entry in json contains id, document and summary. Newline characters and extra space characters are treated in order to ensure smooth processing.

\section{Method}
In this research, in order to attempt to address the issues of cost, time, human efforts in building a data set for low resource language like hindi. It is evident from the previous illustrations that there does not exists a reference for validating the quality of translation.Hence a novel approach of translation and back translation is used to ensure the translation does not lose semantic and context by evaluating against the original source. Three different approaches have been experimented and after studying each approach, the benefits of each approach is amalgamated to form a novel approach to build a Hindi Data set for summarization with least human effort based on the findings from the study. The red color stages depict the S1 System, the Green color depicts the S2 System and Violet color depict the S3 system.
\begin{figure}
    \centering
    \includegraphics[width=1\linewidth]{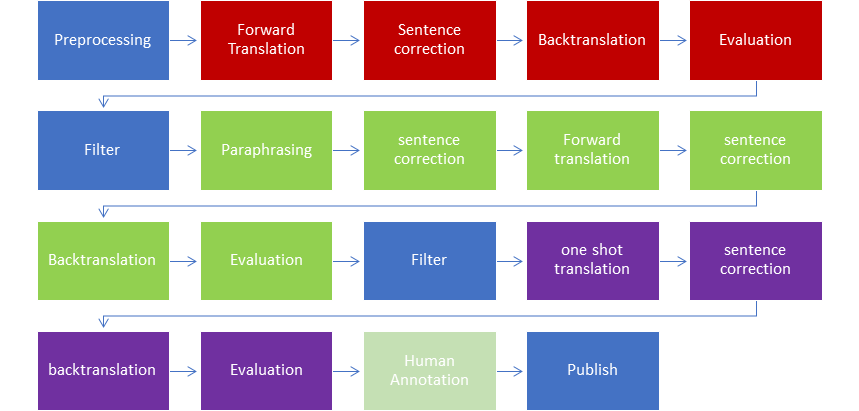}
    \caption{Process Architecture }
    \label{figure1}
\end{figure}

\subsection{Overview}
The process architecture in \ref{figure1}  depicts the process flow of the system that is being proposed. The grouping of the tasks indicates a specific approach to perform the translation. The color coding red, green and purple represent each approach respectively. 
\subsection{Pre processing}
The preprocessing phase is critical as it sets the stage for subsequent translations. This phase may involve text normalization, such as correcting typos, standardizing spelling, removing special characters, or handling casing. Additionally, it may include segmenting the text into sentences or tokens, which can significantly impact the quality of machine translation.
A key decision in this phase is determining the level of normalization required. Over-normalization can strip the text of its nuances, affecting the quality of the translation. Conversely, under-normalization can lead to inconsistencies in the translation process. The choice of tokenization and segmentation strategies can also affect how well the translation tools handle the text, especially considering the syntactic and morphological differences between languages.
\subsection{Forward Translation and Back translation}
 In forward translation, the preprocessed text is converted from the source language into Hindi using a translation tool like LibreTranslate. Although any cost effective language translation tool could be used for this purpose, however it was ensured that both forward translation and backtranslation are performed using same method.  Backtranslation involves translating the Hindi text back to the original language to assess translation quality. 
 The choice of translation tool can greatly affect the outcome. Some tools may be better at handling certain languages or dialects than others. A significant caveat of backtranslation is that a good backtranslation does not necessarily equate to a good forward translation. The quality of the translation is assumed to be high if the backtranslated text closely matches the original, but this might not always be the case due to paraphrasing or different sentence structures that are still correct.
\subsection{NLP Error Correction}
On the translations generated from forward translation  sentence correction methods applied to improve grammatical accuracy and readability. Although any available NER tools can be used for this.In our approach we leveraged the use of google translate to correct a sentence. We provide a hindi text and try to translate it to hindi text which ends up with a corrected formation of the words and phrases in the text. 
Sentence correction tools must also be chosen carefully to ensure they do not introduce new errors or alter the intended meaning. Additionally, the degree of correction needed should be balanced to avoid over-correction, which can sometimes lead to loss of meaning or context.
\subsection{Paraphrasing for retranslation }
In the paraphrasing approach, the aim is to rewrite the text in a way that retains the original meaning but with different phrasing. This step can be particularly useful for generating more natural-sounding translations and avoiding literal translations that may seem awkward.
The paraphrasing step should ensure that the essence of the original text is maintained, which can be challenging. Over-paraphrasing can change the intended meaning or leave out critical information. There is also a decision to be made regarding the degree to which paraphrasing should be automated versus manually reviewed.
\subsection{one shot LLM translation}
This approach leverages a large language model for a one-shot translation, which means directly translating the text into Hindi without iterative corrections. This method relies on the advanced capabilities of large language models to understand and translate context accurately.
The use of a one-shot translation assumes that the language model is well-trained and can handle the nuances of the text without further intervention. This may not always be true, especially for idiomatic expressions or complex syntax. Decisions need to be made regarding the quality thresholds for acceptance and whether additional steps are necessary for sentences that do not meet these thresholds.Another drawback of using this approach is potential model hallucinating. Abstractive summarization is sensitive to hallucination issues. 
\section{Evaluation}
The evaluation phase then uses various metrics like BERTScore, BLEUScore, chrF, chrF++, TER, and COMET to measure the fidelity of the backtranslated text to the original text.

\subsection{Metrics for translation and Backtranslation}
ROUGE Metrics for Text Summarization
ROUGE (Recall-Oriented Understudy for Gisting Evaluation) metrics are widely used to evaluate the quality of summaries by comparing them to one or more reference summaries. The most common ROUGE metrics are:
ROUGE-1 and ROUGE-2: These measure the overlap of unigrams (ROUGE-1) and bigrams (ROUGE-2) between the system-generated summary and the reference summaries. The formulas for recall, precision, and F1-score for ROUGE-N (where N can be 1 or 2) are as follows:

  Recall (R) = \(\frac{\text{No. of overlapping n-grams between system summary and reference summary}}{\text{Total number of n-grams in the reference summary}}\)

  Precision (P) = \(\frac{\text{No. of overlapping n-grams between system summary and reference summary}}{\text{Total number of n-grams in the system summary}}\)

  F1-Score = \(\frac{2 \times \text{Precision} \times \text{Recall}}{\text{Precision + Recall}}\)

ROUGE-L: This measures the longest common subsequence (LCS) between the system-generated summary and the reference summaries. It accounts for sentence-level structure similarity naturally and identifies the longest co-occurring sequence of words. The formula for ROUGE-L's F1-score is derived from the LCS length, considering both precision and recall.

Translation Evaluation Metrics:

CHRF and CHRF++: Character n-gram F-score (CHRF) calculates similarity based on character n-grams, weighting precision and recall. CHRF++ extends this by including word bigrams, enhancing its sensitivity to lexical and morphological variations. The basic form of the CHRF score is a weighted F-score calculated over character n-grams:

  \[ \text{CHRF} = F_{\beta} = (1 + \beta^2) \cdot \frac{\text{Precision} \cdot \text{Recall}}{(\beta^2 \cdot \text{Precision}) + \text{Recall}} \]

  where \(\beta\) is typically set to emphasize recall.

COMET: A neural framework for machine translation evaluation that relies on pre-trained language models to predict the quality of translations based on context. It leverages cross-lingual embeddings and quality estimation models but doesn’t have a simple formula as it involves complex model inference.

BERTScore: Utilizes the embeddings from pre-trained BERT models to compute cosine similarity between the tokens of the system output and reference texts. It computes precision, recall, and F1 measures based on these similarities. The scores are often rescaled to be more interpretable.

\subsection{Metrics for NLP Error correction}
Sentence Error Detection and Correction for Hindi
F1-Score: While specific metrics for Hindi might vary, the F1-score, combining precision and recall, is standard for evaluating error detection and correction tasks across languages, including Hindi. It balances the trade-off between detecting as many errors as possible (recall) and ensuring the detected errors are indeed errors (precision).
\subsection{Metrics for coherence and paraphrasing}
- **BLEU (Bilingual Evaluation Understudy)**: Though originally designed for translation, BLEU can be used for paraphrasing by measuring the n-gram overlap between a candidate paraphrase and one or more reference paraphrases. It emphasizes precision but incorporates a brevity penalty to discourage overly short paraphrases.

Each of these metrics has its strengths and limitations, and the choice of metric often depends on the specific requirements of the evaluation task, the nature of the texts being compared, and the goals of the research or application.
\section{Human Annotation and Publishing}
After automated processes, human annotation serves as a quality assurance step. Skilled linguists review the translations, make corrections, and ensure that the text sounds natural and accurate. The final phase is to publish the dataset, making it available for use by researchers or for integration into applications. In our study we utilize doccano application which provides a convenient way to build a data store and a web UI to annotate and label the translations. The study purpose is to reduce the count of articles that need human annotation. The decision of need for human annotation is done by combination of the TER and BERTScore metrics. 
\begin{figure}
    \centering
    \includegraphics[width=1\linewidth]{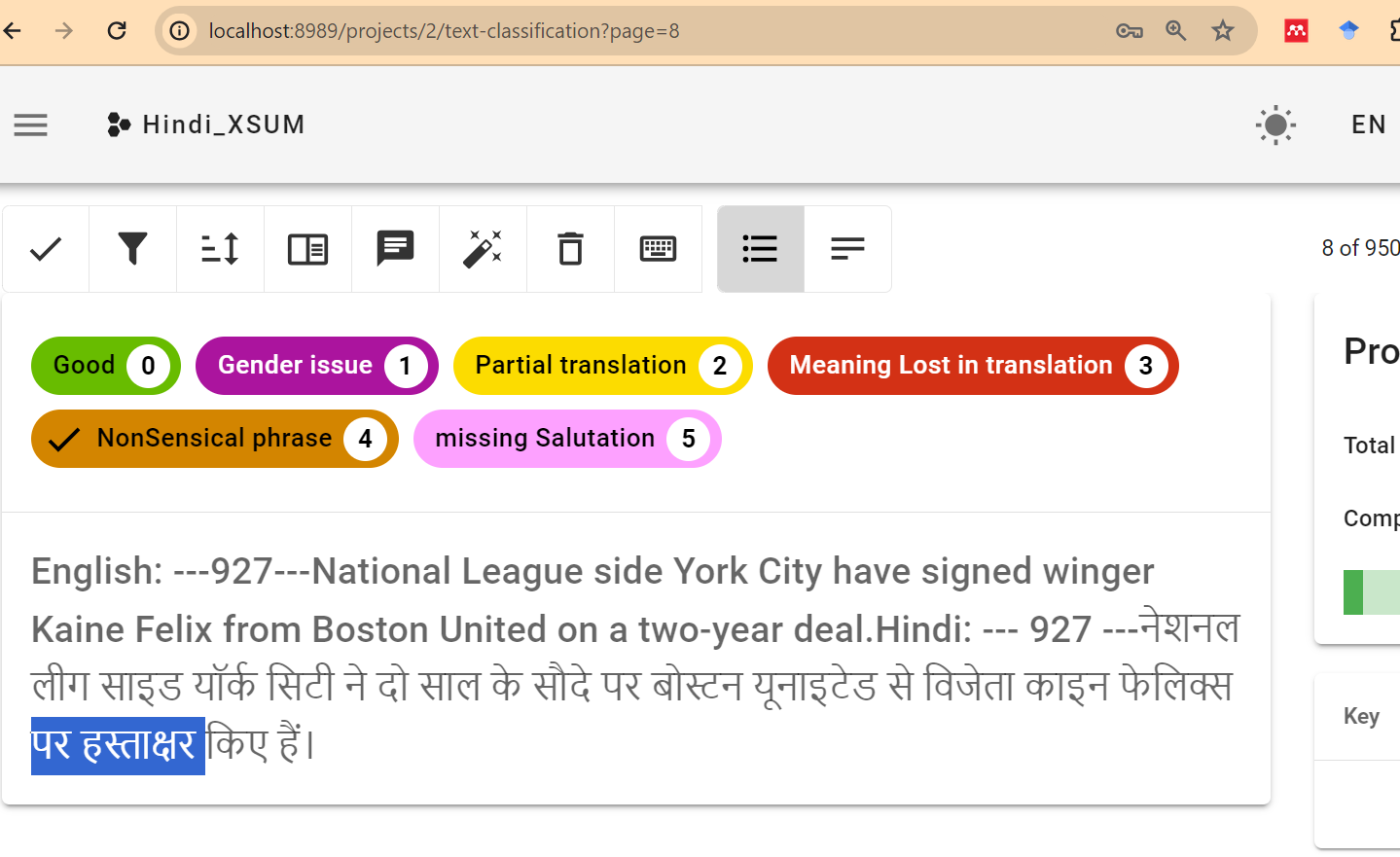}
    \caption{sample Annotation UI used for Human Annotation}
    \label{fig:enter-label}
\end{figure}
Human annotation can be resource-intensive, so it’s important to decide how much text will undergo this scrutiny. There’s also a decision to be made regarding the criteria for selecting texts for human review. Furthermore, deciding on the platform and format for publishing the dataset is crucial for accessibility and ease of use by the intended audience.
\section{Discussion}
After following the process explained. The Evaluation was done after every back translation stage. The evaluation was chosen to be done at this point as it was the only point where a reference data is available to validate, and the goal is automate the evaluation process without any human intervention.  
Table below shows the details of the calculation done during this process.

\begin{table}[ht]
\centering
\begin{tabular}{lrrrrrr}
\hline
Metric & S1 Doc-Min & S1 Doc-Max & S1 Doc-Avg & S1 Sum-Min & S1 Sum-Max & S1 Sum-Avg\\
\hline
bertscore & 0.846509 & 0.946368 & 0.898022 & 0.876450 & 0.986237 & 0.940115 \\
bleu & 0.000000 & 33.783000 & 24.389120 & 0.000000 & 70.398000 & 27.539320 \\
chrf & 52.300168 & 72.903135 & 64.362863 & 46.557201 & 89.837737 & 64.356116 \\
chrfpp & 50.617929 & 67.947679 & 61.428394 & 43.215298 & 88.573273 & 61.647786 \\
ter & 40.659341 & 127.941176 & 70.294538 & 15.384615 & 100.000000 & 52.227001 \\
comet & 0.066845 & 0.740710 & 0.492714 & 0.253312 & 0.972967 & 0.639672 \\
\hline
\end{tabular}
\caption{S1 Document Scores}
\end{table}

\begin{table}[ht]
\centering
\begin{tabular}{lrrrrrr}
\hline
Metric & S2 Doc-Min & S2 Doc-Max & S2 Doc-Avg & S2 Sum-Min & S2 Sum-Max & S2 Sum-Avg\\
\hline
bertscore & 0.813572 & 0.925006 & 0.862327 & 0.848024 & 0.970734 & 0.924662  \\
bleu & 0.869000 & 17.381000 & 6.937600 & 0.000000 & 32.979000 & 10.984320 \\
chrf & 22.409003 & 58.296435 & 45.953035 & 25.941370 & 69.931307 & 47.757525\\
chrfpp & 28.184852 & 54.453105 & 46.027330 & 24.872520 & 66.970590 & 44.580148\\
ter & 60.869565 & 859.322034 & 243.833251 & 46.875000 & 141.666667 & 70.886306\\
comet & -0.194245 & 0.671739 & 0.251626 & -0.178736 & 0.810460 & 0.506541\\
\hline
\end{tabular}
\caption{S2 Document Scores}
\end{table}

\begin{table}[ht]
\centering
\begin{tabular}{lrrrrrr}
\hline
Metric & S3 Doc-Min & S3 Doc-Max & S3 Doc-Avg & S3 Sum-Min & S3 Sum-Max & S3 Sum-Avg\\
\hline
bertscore & 0.841987 & 0.931812 & 0.889471 & 0.871716 & 0.978917 & 0.925248 \\
bleu & 5.865000 & 25.421000 & 17.421800 & 0.000000 & 76.116000 & 19.364440\\
chrf & 39.206734 & 64.759392 & 55.707328 & 36.608171 & 82.230945 & 53.784865\\
chrfpp & 44.837493 & 60.495263 & 53.167102 & 32.105871 & 80.857483 & 50.468500 \\
ter & 53.448276 & 300.000000 & 107.179761 & 8.333333 & 100.000000 & 61.058746\\
comet & 0.163025 & 0.714438 & 0.516124 & -0.101425 & 0.897277 & 0.551339 \\
\hline
\end{tabular}
\caption{S3 Document Scores}
\end{table}
\subsection{Results}
Let's delve into some insights based on the provided data for each metric across the three systems:
1. BERTScore
Insight: S1 shows the highest average BERTScore (0.898), indicating that it produces texts most similar to the reference texts, according to contextual embeddings. S2 has the lowest average score (0.862), and S3 is slightly below S1 with an average of 0.889. The maximum and minimum scores also reflect this trend.
2. BLEU
Insight: The BLEU scores are much lower across all systems compared to BERTScore, highlighting the strictness of BLEU in evaluating literal matches between the generated and reference texts. S1 has the highest average BLEU score (24.389), indicating better literal match quality, while S2 has a significantly lower average (6.938). S3 stands in the middle with an average of 17.422. The minimum scores for S1 and S2 indicate that there are instances where there was no overlap with the reference text at all.
3. CHRF
Insight: CHRF, which focuses on character n-gram matches, shows S1 having the highest average score (64.363), closely followed by S3 (55.707), and S2 having the lowest (45.953). This suggests S1 and S3 are more effective in capturing finer-grained textual similarities than S2.
4. CHRF++
Insight: Similar to CHRF, CHRF++ evaluates character-level similarities but with different weighting. Again, S1 leads in average score (61.428), with S3 (53.167) performing better than S2 (46.027). This reinforces the notion that S1 and S3 are generally more effective at capturing character-level textual nuances.
5. TER
Insight: TER (Translation Edit Rate) scores show a wide range across systems, with S1 having a relatively lower average score (70.295), indicating fewer edits needed to match the reference text. S2, however, shows a significantly higher average (243.833), suggesting a higher degree of discrepancy between generated and reference texts. S3's average (107.18) suggests moderate performance. Notably, S2's max TER is extraordinarily high (859.322), indicating an outlier or an extremely poor match in at least one instance.
6. COMET
Insight: COMET scores, which also evaluate semantic similarity, show S1 and S3 with similar average performances (0.493 and 0.516, respectively), both outperforming S2 (0.252). The negative minimum score in S2 (-0.194) indicates instances where the generated text was semantically very different from the reference.
Overall Insights
S1 tends to perform consistently well across all metrics, suggesting its generated texts are both literally and semantically closer to the reference texts.
S2 generally shows lower performance, especially in literal matching (as indicated by BLEU and TER) and semantic alignment (as reflected by its COMET scores).
S3 often ranks between S1 and S2, suggesting a balanced but slightly less effective approach than S1 in terms of matching the reference texts both literally and semantically.

\begin{figure}[ht]
    \begin{minipage}{0.48\textwidth}
        \centering
        \includegraphics[width=\linewidth]{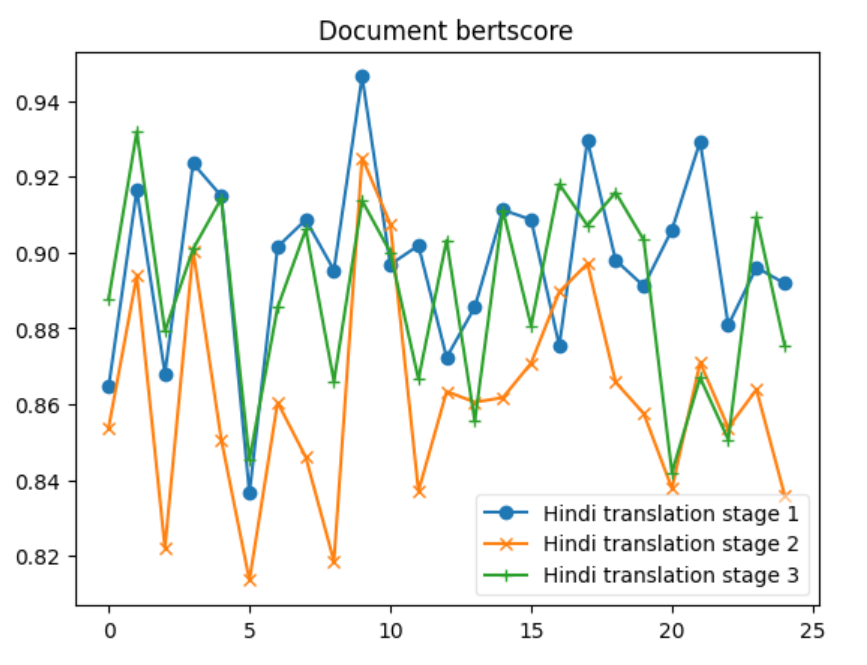}
        \caption{Document BertScore}
        \label{fig:docbertscore}
    \end{minipage}
    \hfill
    \begin{minipage}{0.48\textwidth}
        \centering
        \includegraphics[width=\linewidth]{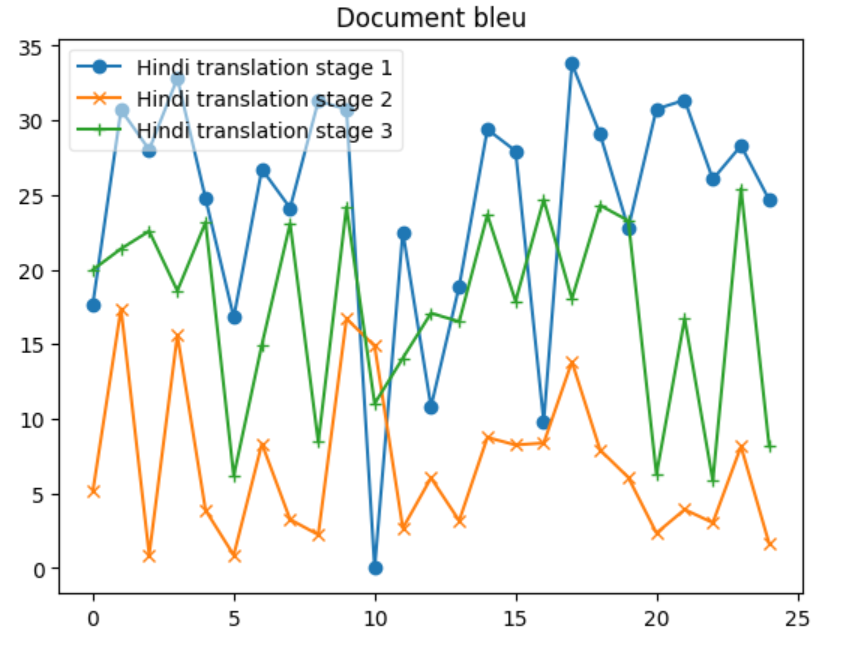}
        \caption{Document BLEU Score}
        \label{fig:docbleuscore}
    \end{minipage}
\end{figure}

\begin{figure}[ht]
    \begin{minipage}{0.48\textwidth}
        \centering
        \includegraphics[width=\linewidth]{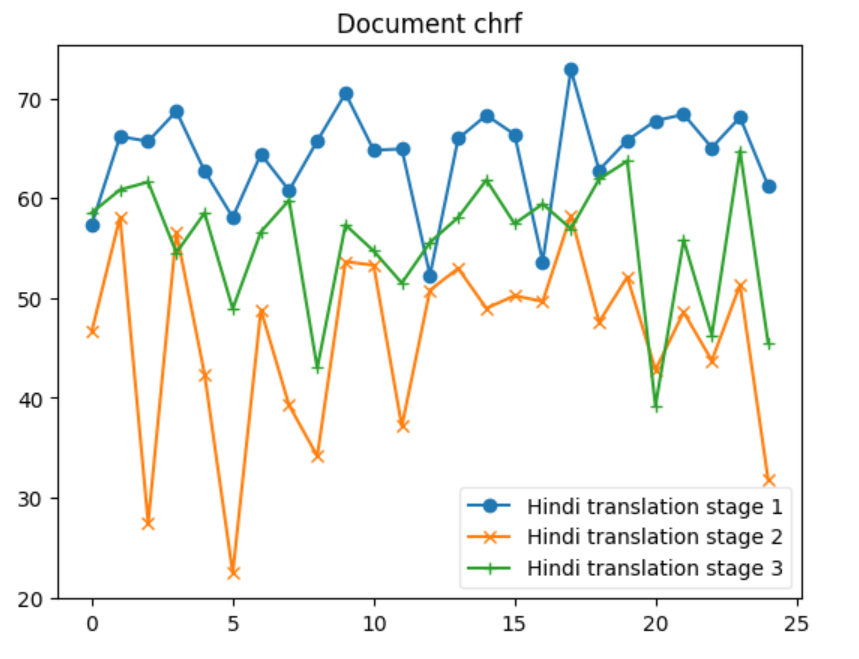}
        \caption{Document CHRF score}
        \label{fig:docchrf}
    \end{minipage}
    \hfill
    \begin{minipage}{0.48\textwidth}
        \centering
        \includegraphics[width=\linewidth]{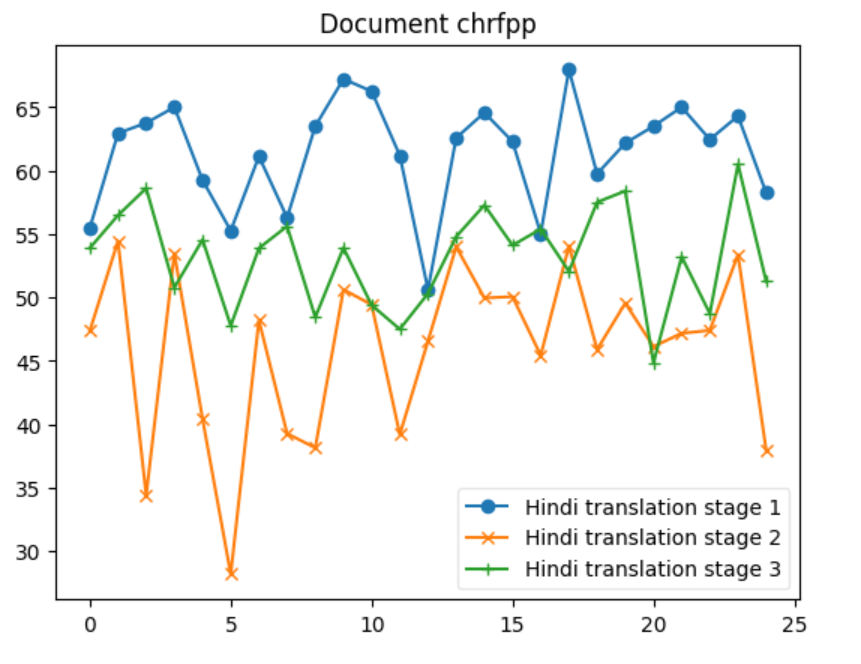}
        \caption{Document CHRF++ score}
        \label{fig:docchrfpp}
    \end{minipage}
\end{figure}

\begin{figure}[ht]
    \begin{minipage}{0.48\textwidth}
        \centering
        \includegraphics[width=\linewidth]{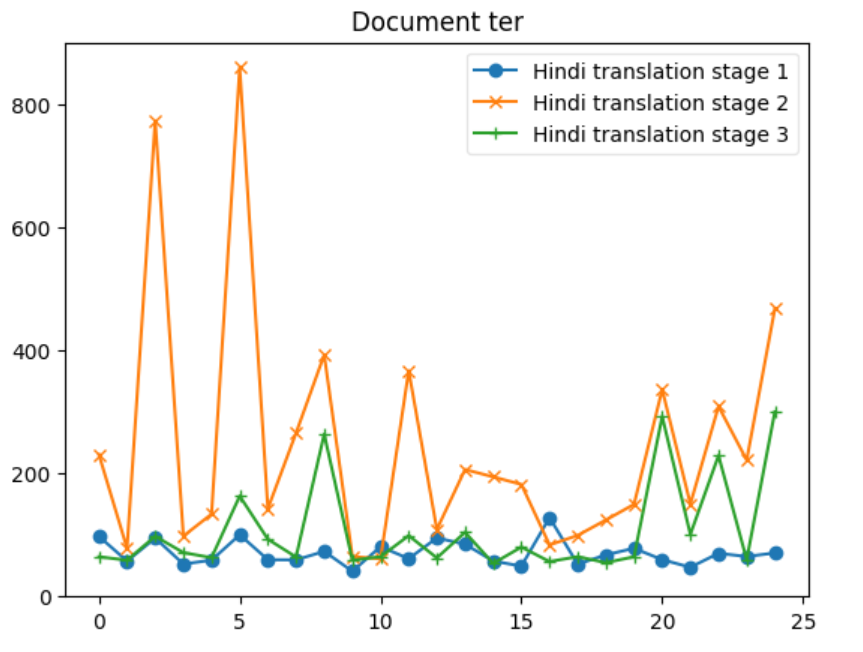}
        \caption{Document TER score}
        \label{fig:docter}
    \end{minipage}
    \hfill
    \begin{minipage}{0.48\textwidth}
        \centering
        \includegraphics[width=\linewidth]{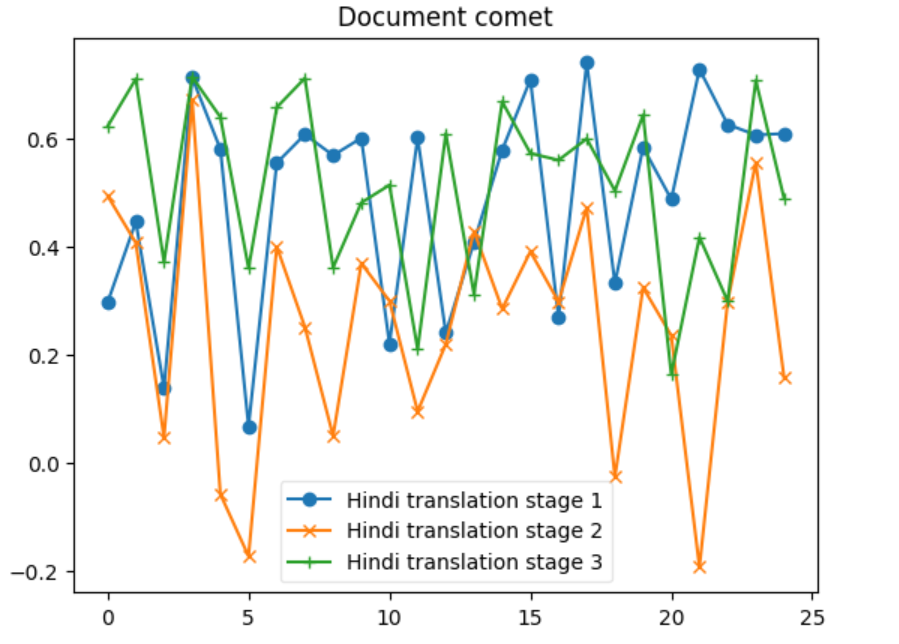}
        \caption{Document COMET score}
        \label{fig:doccomet}
    \end{minipage}
\end{figure}
\begin{figure}[ht]
    \centering
    \begin{minipage}{0.45\linewidth}
        \centering
        \includegraphics[width=\linewidth]{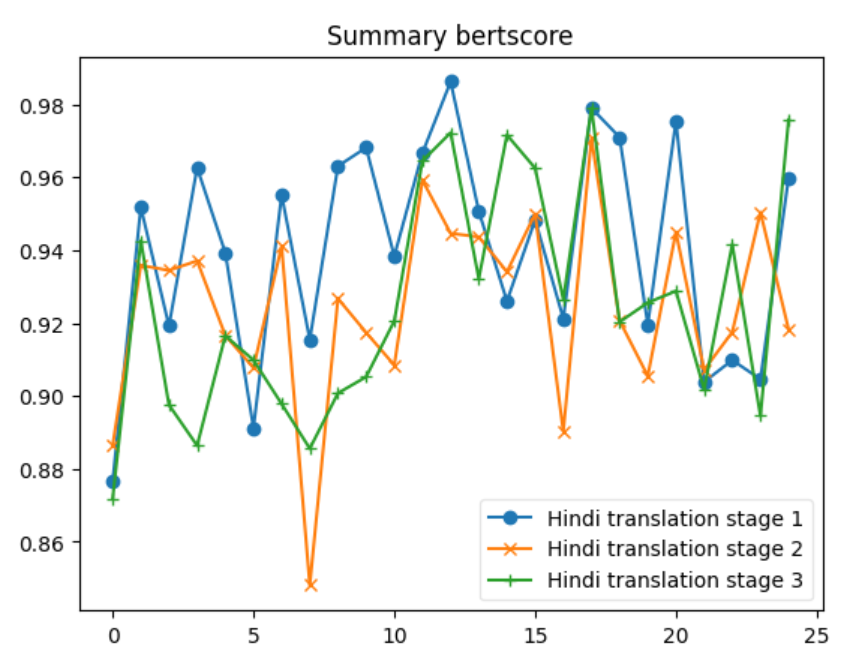}
        \caption{Summary BERTScore}
        \label{fig:summary_bertscore}
    \end{minipage}
    \hfill
    \begin{minipage}{0.45\linewidth}
        \centering
        \includegraphics[width=\linewidth]{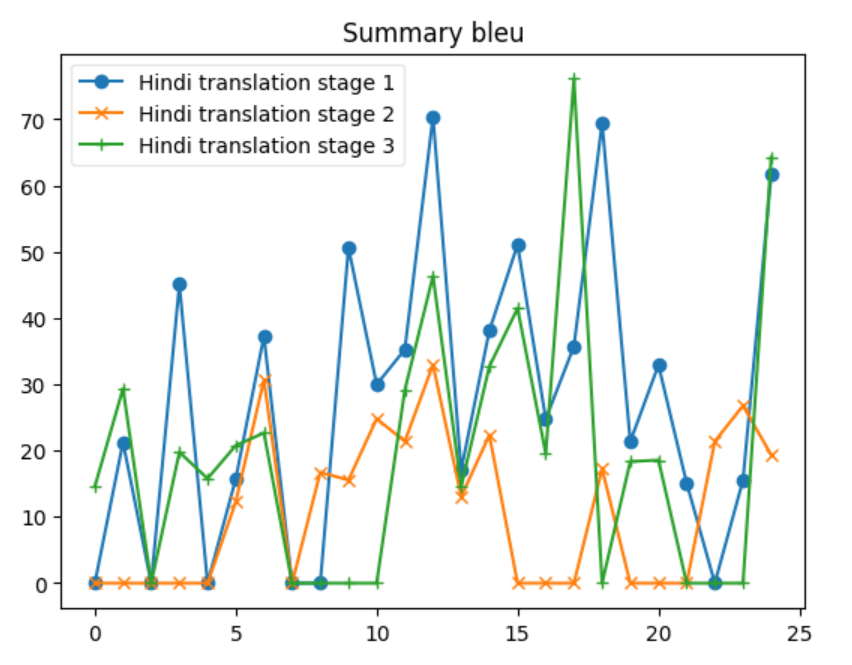}
        \caption{Summary BLEU Score}
        \label{fig:summary_bleu}
    \end{minipage}
\end{figure}

\begin{figure}[ht]
    \centering
    \begin{minipage}{0.45\linewidth}
        \centering
        \includegraphics[width=\linewidth]{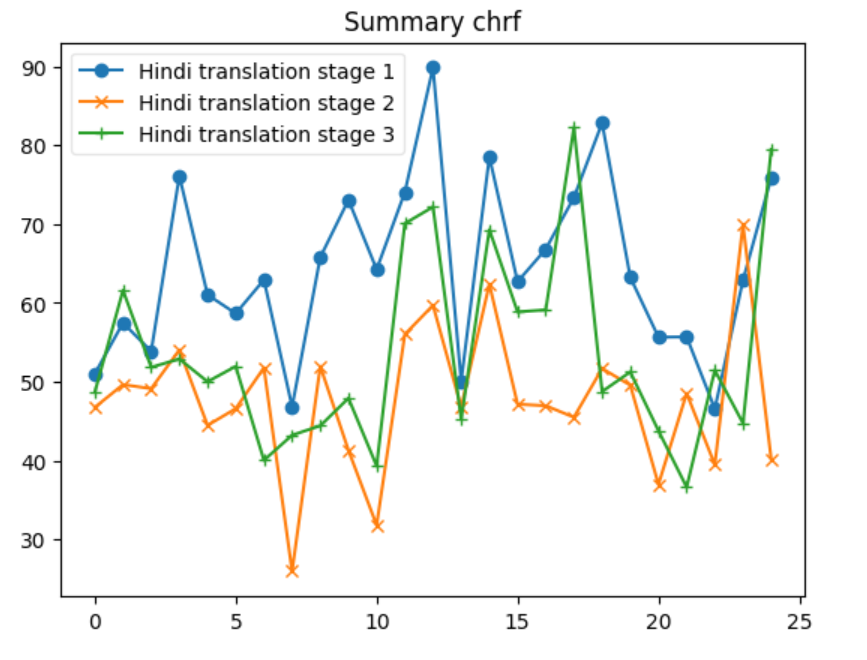}
        \caption{Summary CHRF Score}
        \label{fig:summary_chrf}
    \end{minipage}
    \hfill
    \begin{minipage}{0.45\linewidth}
        \centering
        \includegraphics[width=\linewidth]{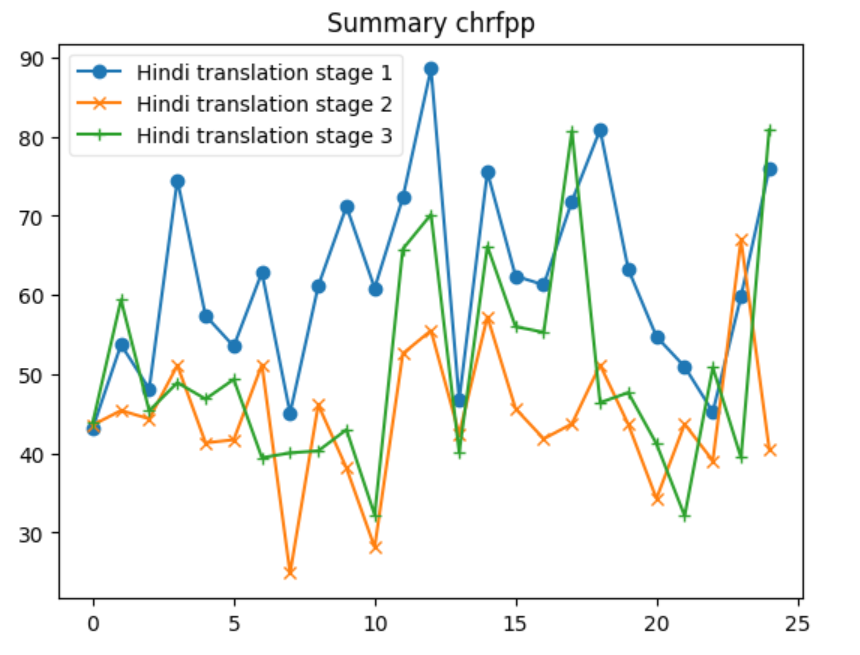}
        \caption{Summary CHRF++ Score}
        \label{fig:summary_chrfpp}
    \end{minipage}
\end{figure}

\begin{figure}[ht]
    \centering
    \begin{minipage}{0.45\linewidth}
        \centering
        \includegraphics[width=\linewidth]{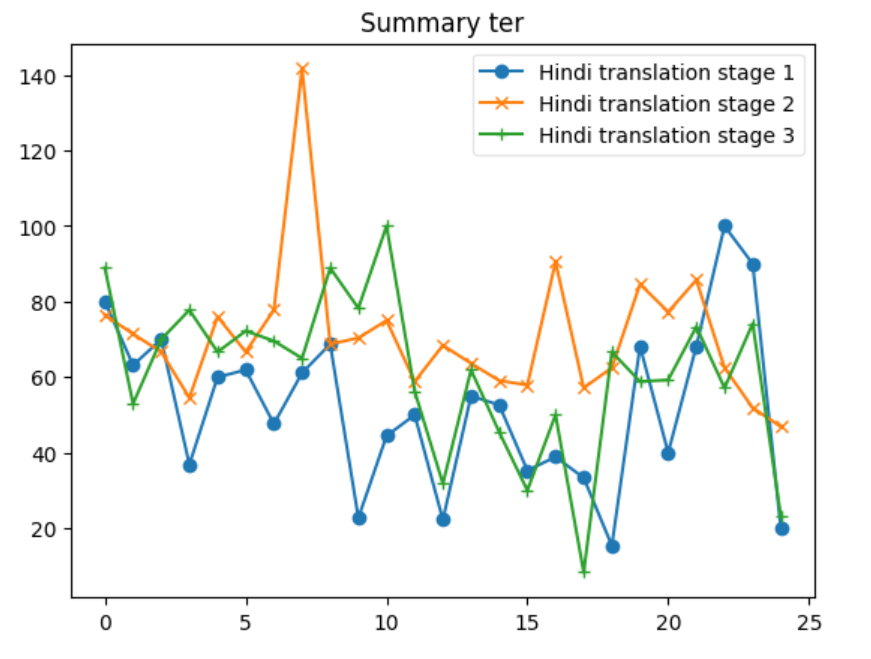}
        \caption{Summary TER Score}
        \label{fig:summary_ter}
    \end{minipage}
    \hfill
    \begin{minipage}{0.45\linewidth}
        \centering
        \includegraphics[width=\linewidth]{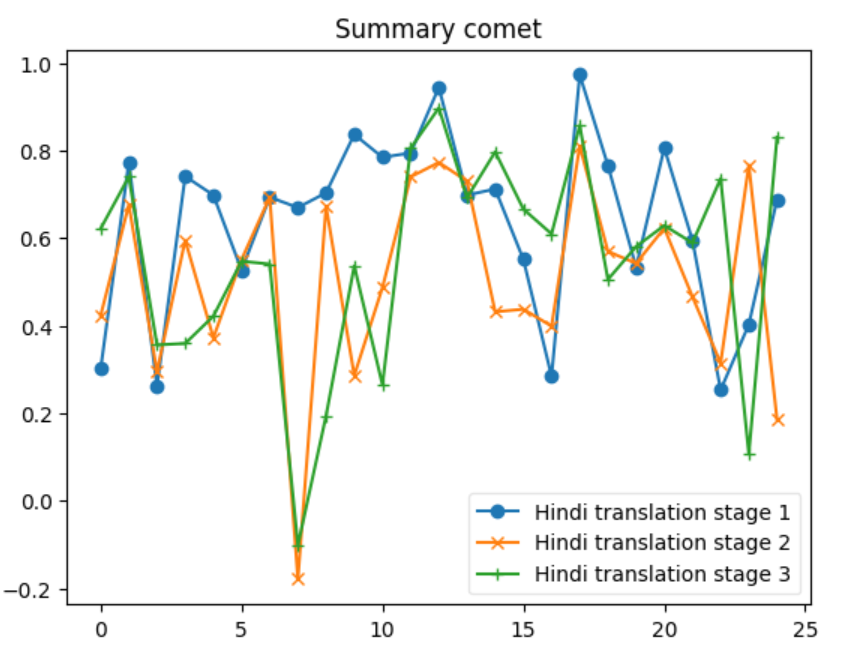}
        \caption{Summary COMET Score}
        \label{fig:summary_comet}
    \end{minipage}
\end{figure}

In our analysis, we discern that systems employing sequence-to-sequence translation methodologies, exemplified by Google Translate and LibreTranslate (S1 system), surpass alternative approaches in terms of performance. The appeal of these systems lies in their continuous enhancement and unrestricted accessibility. Conversely, S3 systems, which utilize large language models (LLMs) for one-shot translation, demonstrate competitive performance relative to S1 systems. However, their adoption is hindered by substantial costs, as pricing is contingent upon the number of tokens processed. Meanwhile, methods based on paraphrasing present a more economical option, yet they suffer from significant variability in their performance.

Additionally we also looked ahead to validate if the relationship between the original document and Summary from XSUM is being maintained in the translation Data set. Both the below graphs depict instances where the F1-scores for the Hindi Translation reach or exceed the scores of the Original Dataset. This suggests that for several entries, the translated summary text performed similar and was able to retain a substantial amount of the semantic content from the original texts. 

\begin{figure}
    \centering
    \includegraphics[width=0.5\linewidth]{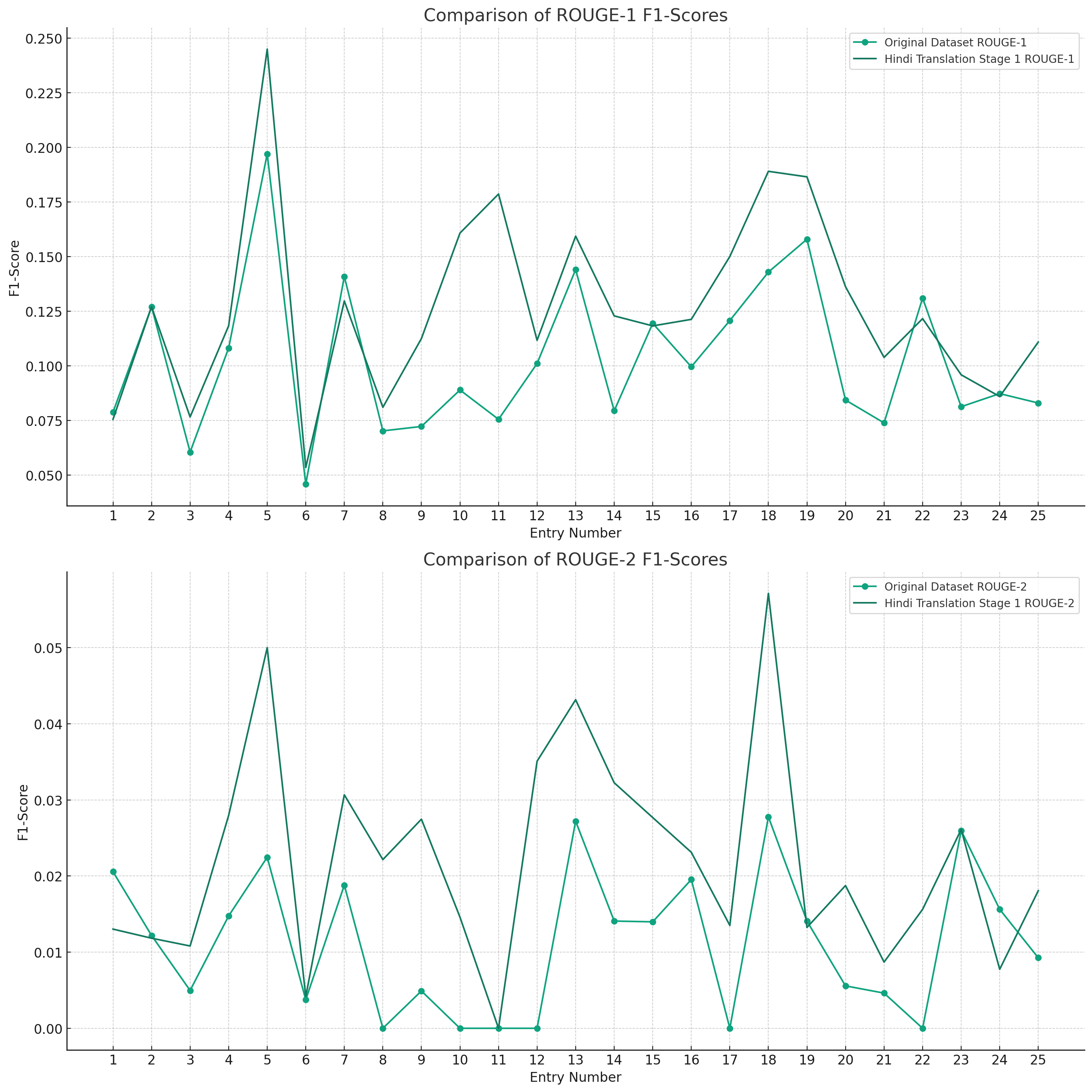}
    \caption{Comparison of  English XSUM and Hindi Translated XSUM  over Rouge-1and Rouge-2 evaluation against their corresponding Original Document }
    \label{fig: R1R2_compare  }
\end{figure}
\begin{figure}
    \centering
    \includegraphics[width=0.5\linewidth]{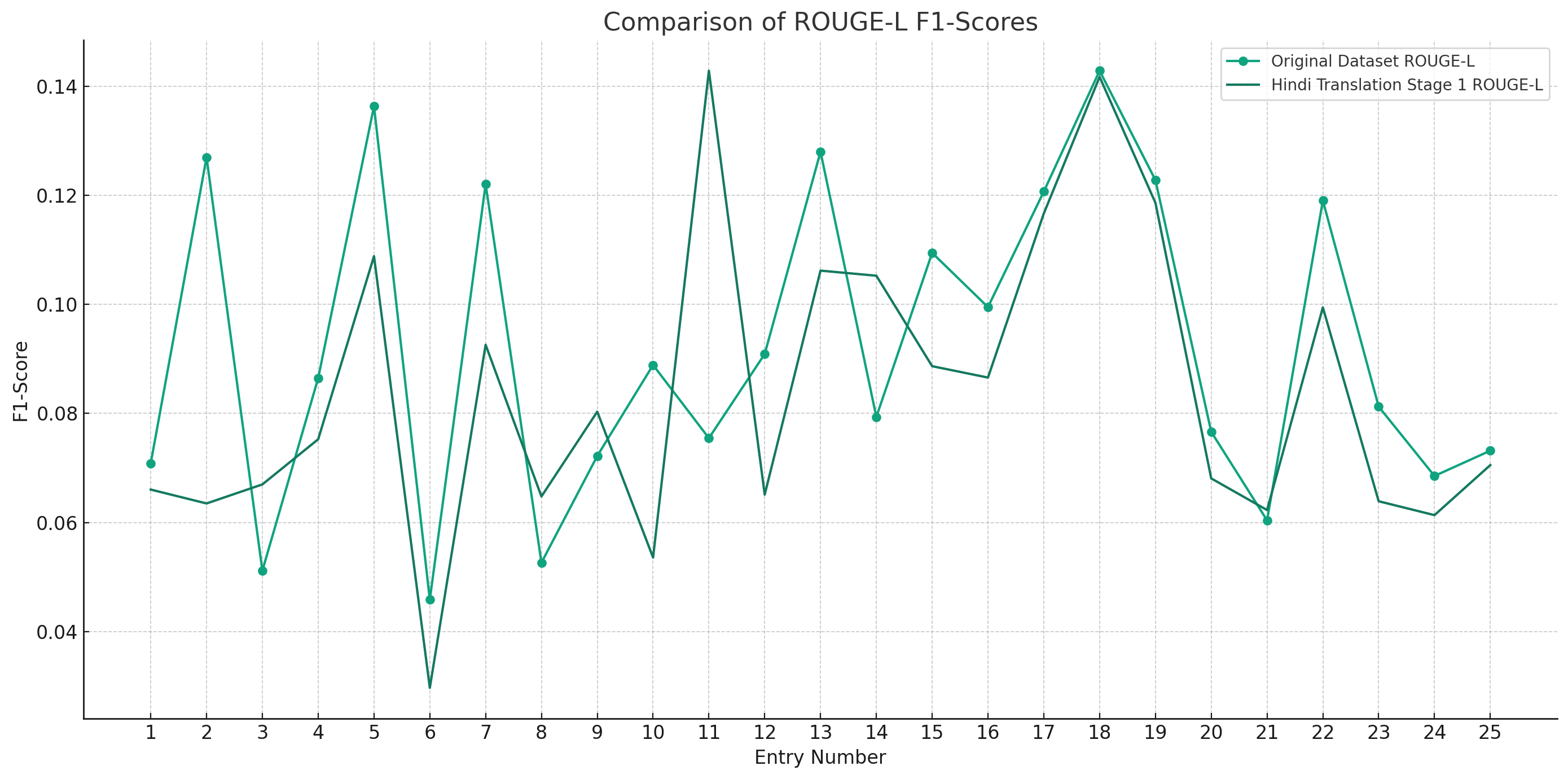}
    \caption{Comparison of English XSUM and Hindi Trnaslated XSUM over ROUGE-L evaluation against their corresponding Original Document}
    \label{fig:RL_compare}
\end{figure}
The patterns observed across the different ROUGE metrics show that there is a level of consistency in the translation process. Despite some dips, the general trend indicates that the Hindi translations are quite often on par with the original text, which is a commendable aspect of the translation process. 

In light of these observations, our research proposes a novel strategy that sequentially integrates S1, S2, and S3 systems. This approach aims to preliminarily filter articles requiring translation through their Translation Edit Rate (TER) and BERT scores. Consequently, this methodology significantly reduces the necessity for translations via LLM-based approaches, thereby facilitating the generation of a dataset through a markedly more cost-effective process.The translated dataset has been publicly shared on the Hugging Face forum under \href{https://huggingface.co/pkumark/Hindi_XSUM}{pkumark/Hindi\_XSUM}.

\section{Future scope}
\subsection{Enhancing Translation Through Automated Systems}

In the pursuit of optimizing translation processes, our approach leverages a multifaceted strategy that incorporates automated metrics, capitalizes on resources abundant in high-resource languages, and employs auto-correction methods to enhance translation quality. These components collectively contribute to a more efficient and accurate translation methodology suitable for inclusion in a wide array of research contexts.

Automated Metrics for Comprehensive Evaluation
At the core of our methodology lies the utilization of automated metrics, such as Translation Edit Rate (TER) and BERT scores, to evaluate various dimensions of translation quality. These metrics offer a systematic and objective means to assess the accuracy, fluency, and semantic coherence of translations across different systems. By automating the evaluation process, we significantly reduce the time and effort required to analyze translation outputs, enabling a rapid comparison of different translation approaches. This automation facilitates the identification of the most effective translation strategies, ensuring that our selection of tools and methods is grounded in empirical evidence.

Leveraging High-Resource Language Capabilities
Another pivotal aspect of our approach is the strategic use of resources available for high-resource languages, particularly English. High-resource languages benefit from a wealth of linguistic data, advanced natural language processing tools, and extensive research contributions. By tapping into these resources, we enhance our translation framework's ability to handle complex linguistic structures and idiomatic expressions, thus improving the quality of translations from and into less-resourced languages. This approach not only broadens the applicability of our translation methodology but also contributes to bridging the resource gap between languages.

Auto-Correction and Transliteration for Refined Outputs
Furthermore, our approach incorporates automatic sentence correction methods to address immediate translation errors and perform transliteration of noun components. This step is crucial for refining the initial translations by correcting grammatical inaccuracies and adapting proper nouns to the target language's orthographic system. The auto-correction process ensures that translations are not only semantically accurate but also adhere to the grammatical conventions of the target language. Simultaneously, transliteration mechanisms maintain the integrity of names and specialized terms, preserving their recognizability across linguistic contexts. These enhancements are instrumental in achieving translations that are both accurate and culturally resonant.

In conclusion, by integrating automated evaluation metrics, leveraging resources for high-resource languages, and employing advanced correction techniques, our approach presents a comprehensive solution to the challenges of translation. This methodology promises to deliver high-quality translations efficiently and cost-effectively, marking a significant advancement in the field of automated translation.

\subsection{Scalability to Other Low-Resource Languages}

A notable strength of the translation methodology delineated in this research is its inherent adaptability and scalability to languages beyond Hindi, particularly those classified as low-resource. This versatility is foundational to our approach, designed with the flexibility to accommodate and enhance translation processes for languages that traditionally lack extensive linguistic datasets and advanced natural language processing (NLP) tools.

Framework Adaptability
The framework's reliance on automated metrics for evaluating translation quality is language-agnostic. Metrics such as Translation Edit Rate (TER) and BERT scores can be applied universally, regardless of the language in question. This universality allows for the objective assessment of translation outputs across a broad spectrum of languages, facilitating the identification and adoption of the most effective translation techniques tailored to each language's unique characteristics.

Leveraging Cross-Linguistic Resources
Our methodology's emphasis on utilizing resources from high-resource languages serves as a significant advantage when extending the approach to other low-resource languages. The techniques developed and lessons learned from translating between English (or another high-resource language) and Hindi can be applied to similar translation challenges faced by other low-resource languages. This transfer of knowledge and resources fosters a collaborative environment that elevates the translation capabilities for a wide range of languages, contributing to a more inclusive global digital landscape.

Auto-Correction and Transliteration Across Languages
The integration of auto-correction and transliteration methods in our approach is designed to be adaptable to the linguistic intricacies of different languages. These techniques are critical for refining translation outputs and ensuring the preservation of cultural and contextual integrity across languages. By customizing these methods to address the grammatical and orthographic nuances of each target language, our approach ensures that translations are not only accurate but also culturally appropriate.

Bridging the Resource Gap
The application of this comprehensive translation methodology to other low-resource languages has the potential to significantly bridge the gap in linguistic resources and NLP tools available to these languages. By enhancing the quality and efficiency of translations, we contribute to the broader goal of linguistic equity, ensuring that speakers of low-resource languages have access to information, technology, and communication platforms on par with high-resource language communities.

In summary, the translation methodology developed in this research extends beyond its immediate application to Hindi, offering a scalable and adaptable framework for improving translation processes for a variety of low-resource languages. This adaptability not only enriches the linguistic diversity of digital content but also empowers communities around the globe by providing them with access to accurate and culturally resonant translations.
\section{Conclusion}
This research delineates an innovative translation methodology that harmonizes automated evaluation metrics, the extensive resources of high-resource languages, and sophisticated auto-correction techniques, demonstrating a significant leap in the translation quality of low-resource languages, with an initial focus on Hindi. Our approach showcases the utility of sequence-to-sequence models and the strategic application of large language models (LLMs), balancing performance with cost-effectiveness through the selective use of advanced translation techniques.

Key to our findings is the potential scalability of this methodology to other low-resource languages, promising a reduction in the linguistic resource gap and advancing global communication inclusivity. The methodology's success in leveraging automated metrics for preliminary content filtering illustrates a path toward efficient and accessible translation processes, suggesting a sustainable model for enhancing linguistic diversity in the digital age.

In essence, this work contributes to a more linguistically equitable digital environment, fostering enhanced access to information and technology across language barriers. It lays the groundwork for future innovations in translation technology, aiming for a world where language diversity is not a barrier but a bridge to richer global understanding and cooperation.
\appendix

\chapter{Appendix}

\section{Appendix Section}

Nothing doing\ldots

\bibliography{sample}
\printindex
\end{document}